\documentclass[review]{elsarticle}

\usepackage[left = 3cm,
        right=3cm,
        top=3cm,
        bottom=3.5cm
]{geometry}

\def\phcomments{1}
\newcounter{note}[section]
\renewcommand{\thenote}{\thesection.\arabic{note}}
\newcommand{\note}[2]{\refstepcounter{note}\marginpar{\small\bf \textcolor{red}{#1~\thenote}}$\ll${\sf \textcolor{red}{#1's
 Comment~\thenote:}} {\em \textcolor{red}{#2}}$\gg$}

\ifnum\phcomments=1
\newcommand{\notePH}[1]{\note{PH}{#1}}
\else
\newcommand{\notePH}[1]{}
\fi

\ifnum\phcomments=1
\newcommand{\noteBQ}[1]{\note{BQ}{#1}}
\else
\newcommand{\noteBQ}[1]{}
\fi

\usepackage{soul}
\usepackage{lineno,hyperref}
\usepackage{amsfonts}
\usepackage{amsmath}
\usepackage{graphicx}
\usepackage[dvipsnames]{xcolor}
\usepackage{ulem}
\usepackage{booktabs,dcolumn,caption}
\usepackage{graphics}
\usepackage{placeins}
\usepackage{subcaption}

\biboptions{sort&compress}

\usepackage{amssymb}

\newcommand{\Xb}{\mathbf{X}}
\newcommand{\Yb}{\mathbf{Y}}
\newcommand{\Wb}{\mathbf{W}}

\newcommand{\Qb}{\mathbf{Q}}

\newcommand{\Kb}{\mathbf{K}}
\newcommand{\Vb}{\mathbf{V}}

\newcommand{\xb}{\mathbf{x}}
\newcommand{\yb}{\mathbf{y}}

\modulolinenumbers[5]

\journal{International Journal of Forecasting}

\begin{document}

\begin{frontmatter}

\title{Inter-Series Transformer: Attending to Products in Time Series Forecasting}


\author[mit]{Rares Cristian}
\author[ibm]{Pavithra Harsha}
\author[mit]{Clemente Ocejo}
\author[mit]{Georgia Perakis}
\author[ibm]{Brian Quanz}
\author[mit]{Ioannis Spantidakis}
\author[mit]{Hamza Zerhouni}
\affiliation[ibm]{organization={IBM Research},
            addressline={IBM TJ Watson Research Center},
            city={Yorktown Heights},
            postcode={10598},
            state={NY},
            country={USA}}
\affiliation[mit]{organization={MIT},
            addressline={77 Massachusetts Ave},
            city={Cambridge},
            postcode={02139},
            state={MA},
            country={USA}}

\begin{abstract}
Time series forecasting is an important task in many fields ranging from supply chain management to weather forecasting. Recently, Transformer neural network architectures have shown promising results in forecasting on common time series benchmark datasets. However, application to supply chain demand forecasting, which can have challenging characteristics such as sparsity and cross-series effects, has been limited.


In this work, we explore the application of Transformer-based models to supply chain demand forecasting.  In particular, we develop a new Transformer-based forecasting approach using a shared, multi-task per-time series network with an initial component applying attention across time series, to capture interactions and help address sparsity.
We provide a case study applying our approach to successfully improve demand prediction for a medical device manufacturing company.
To further validate our approach, we also apply it to public demand forecasting datasets as well and demonstrate competitive to superior performance compared to a variety of baseline and state-of-the-art forecast methods across the private and public datasets.
\end{abstract}

\end{frontmatter}


\section{Introduction}

\noindent Time series forecasting is a fundamental problem in machine learning with applications across many domains. Common applications of time series forecasting include supply chain management \cite{10.2307/20202189}, financial modeling \cite{KIM2003307}, weather forecasting \cite{chatfield2000time}, and many more. Since many of these problems have been around for far longer than more modern problems in machine learning such as robotic manipulation, often the models used in production are more theoretical in foundation and rely less heavily on data.  That is, they often encode strong priors, or inductive biases \cite{mitchell1980need,gordon1995evaluation}, from the assumptions made and the corresponding simple parametric form the models take. Furthermore, they may not be designed to fully utilize the broader set of related data often available these days for forecasting particular time series.  Traditional time series forecasting methods like exponential smoothing \cite{mckenzie1984general}, state space models \cite{hyndman2008forecasting}, and  auto-regressive ARIMA models \cite{box1968some,hyndman2018forecasting} are still widely used in industry.

As our world becomes increasingly connected and availability of data rises, so does the pursuit of deep learning models (e.g., deep neural networks) to tackle complex tasks such as image identification or voice recognition systems, as well as modeling sequential data such as natural language and time series.
Recurrent neural networks (RNNs) \cite{rumelhart1986learning,lipton2015critical} were one of the first types of neural network model architectures developed for sequential data, and variants such as long-short term memory (LSTM) \cite{hochreiter1997long} have also been popularly used in forecasting \cite{gers2001applying,salinas2020deepar,bandara2020,lai2018modeling,salinas2019high,smyl2020hybrid,nguyen2021temporal}.  However, the recurrent and state-based process of RNNs limits the amount of historical information retained and used in predictions and can also make training more challenging. One family of models that takes inspiration from RNNs and also aims to solve many of its limitations via leveraging attention instead of recurrence to model sequential data is the Transformer model \cite{vaswani2017attention} family.  Transformer models have shown promising results for modeling sequential data across many domains  \cite{lin2022survey}, including recently in time series forecasting \cite{wen2022Transformers,li2019enhancing,wu2020adversarial,lim2021temporal,tang2021probabilistic,lin2021ssdnet,zhou2021informer,wu2021autoformer,liu2021pyraformer,shen2022tcct,cirstea2022triformer,chen2022learning,drouin22tactis,zhou2022fedformer,liu2022non}. These Transformer forecast models are all general-purpose by design and results are reported on a variety of common benchmark datasets, often with clear temporal patterns and signals. However, their application to supply chain demand forecasting has been limited, with only the Temporal Fusion Transformer (TFT) \cite{lim2021temporal} paper including a retail dataset in results, and without these models being specialized for demand forecasting. Supply chain demand time series can exhibit challenging properties, such as sparsity in terms of sales observations at the granular product-location level and skewed value distributions (as it is count data), cross-product effects (demand / sales change for one product can affect demand for others) \cite{gelper2016,leeflang2008,srinivasan2005identifying}, and misalignment / changing sets of time series as new products are added and removed over time, which may not be fully addressed via existing Transformer models.  Furthermore, some recent work called into question the benefits of Transformers for forecasting by showing much simpler models were often able to outperform a large set of recently proposed Transformer forecast models on the same common benchmark datasets \cite{zeng2022Transformers}. This highlights the need for more study of Transformer model application on targeted real world datasets for particular tasks. To enhance the practical application of Transformers in supply chain demand forecasting and potentially supplant the use of classical models, it is crucial to conduct more case studies that report results and analyses of applying Transformers on real-world data, which is currently lacking.


In this work we aim to explore the application of Transformer models to supply chain demand forecasting.  We make the following contributions:
\begin{itemize}
    \item We develop new Transformer model variations targeting supply chain demand forecasting to address the aforementioned challenging characteristics.  This includes developing a new architecture consisting of an initial component applying self-attention transformation across time series followed by a shared per-time series Transformer network, applied independently per time series with shared training across time series in a multi-task manner.   This can enable capturing descriptive interactions between time series via the initial cross-series attention transformation while also helping to address sparsity in the form of limited observations both in terms of limited individual time series length and limited non-zero demand. The multi-task aspect (shared network applied to each time series) can also help overcome overfitting associated with a direct multivariate approach and along with the initial cross-series attention allows for application when the set of time series changes over time.  In short our approach is aimed at achieving the best of both worlds (multivariate / cross-series modeling and multi-task per-time series modeling) compared to prior approaches.
    \item We provide a case study applying Transformers and our approach to forecast demand for a medical device manufacturing company, analyzing results for multiple horizons and target metrics to support their different business needs. We successfully demonstrate that our method significantly reduces forecast errors compared to their currently used classical forecasting method as well as many other other time series models including a variety of neural network and Transformer forecasters.
    \item We compare our proposed approach on two additional publicly available retail demand forecasting datasets to give a better sense of performance and how it compares to other methods. We show competitive to superior performance on two publicly available retail datasets compared to Transformer and state-of-the-art baselines.
    \item Finally, we further present additional analyses and an ablation study, examining which aspects of our model were advantageous and which did not enhance Transformer forecasting in this case study.
\end{itemize}

In summary our new approach is specifically designed and evaluated to better tackle the challenges of supply chain demand forecasting such as sparsity, overfitting and cross-series effects. We focus particularly on demonstrating the effectiveness of our methodology through a novel case study aimed at improving demand prediction for a medical device manufacturing company, as well as additional demand forecasting data sets.



\section{Related Work}
\label{Related Work}

Time series forecasting has emerged in recent years as a crucial topic in machine learning. There has been a growing interest in using time series models for forecasting in various domains such as transportation \cite{salinas2020deepar, bandara2020, lai2018modeling, salinas2019high, nguyen2021temporal, li2019enhancing, wu2020adversarial, lim2021temporal, tang2021probabilistic, wu2021autoformer, shen2022tcct, zhou2022fedformer, liu2022non}, energy \cite{lai2018modeling, salinas2019high, nguyen2021temporal, li2019enhancing, wu2020adversarial, tang2021probabilistic, lin2021ssdnet, zhou2021informer, wu2021autoformer, shen2022tcct, cirstea2022triformer, chen2022learning, zhou2022fedformer, liu2022non}, weather \cite{zhou2021informer, wu2021autoformer, cirstea2022triformer, chen2022learning, zhou2022fedformer, liu2022non} or retail sales \cite{salinas2020deepar}. As evidenced by the recent M5 competition \cite{MAKRIDAKIS20221325}, there is also growing interest in developing new machine learning and deep learning time series methods for supply chain demand forecasting specifically. However, the literature on this kind of time series models in the context of retail demand forecasting is limited. We provide a brief overview of existing work to establish the motivation and context for our proposed approach.

\subsection{Traditional time series models}
Many traditional time series forecasting methods are still used in industry \cite{mckenzie1984general, hyndman2008forecasting, box1968some, hyndman2018forecasting}. Exponential smoothing \cite{mckenzie1984general, hyndman2008forecasting} produces a prediction that is a weighted sum of past observations with exponentially decreasing weights for past data points. Holt-Winters \cite{hyndman2008forecasting}, that falls into the exponential smoothing family, adds model components to enable capturing trend and seasonality. Autoregressive Integrated Moving Average (ARIMA) \cite{box1968some, hyndman2018forecasting} is a large class of models combining an auto-regressive model, a moving average model and a differencing step to forecast stationary and non-stationary time series.  As we will see later, these traditional methods often have limitations compared to machine learning and deep learning time series models, especially when it comes to large and complex datasets. In the framework of the M5 competition \cite{MAKRIDAKIS20221325}, the results achieved show that the traditional time series methods were outperformed by state-of-the-art machine learning models.

The book \cite{hyndman2018forecasting}  provides a comprehensive reference to a range of traditional time series forecasting methods including exponential smoothing, ARIMA and state space models.

\subsection{RNN and CNN-based time series models}
A Recurrent Neural Network (RNN)
is a type of neural network used for sequential data with the ability of capturing past information stored by having a recurrent hidden state whose activation at each time is dependent on the previous time state and current time input. RNN-based models can be used for language translation, natural language processing or image identification as well as for time series forecasting showing promising results in this area. RNN-based models have been extensively applied to time series forecasting \cite{rumelhart1986learning,lipton2015critical, hochreiter1997long, gers2001applying,salinas2020deepar,bandara2020,lai2018modeling,salinas2019high,smyl2020hybrid,nguyen2021temporal}. One variant of Recurrent Neural Network is the Long-Short Term Memory model (LSTM) \cite{lipton2015critical, hochreiter1997long}. The purpose of this method is to overcome the challenge of long-term dependencies in time series forecasting by introducing a sophisticated hidden layer that controls the flow of information required to predict the output in the network. By maintaining a memory of past information, the hidden layer regulates the flow of data and stores the relevant information to use for the next steps. This method can capture long-term patterns in the time series and make more accurate predictions for future time points.
 DeepAR \cite{salinas2020deepar} is a forecasting method based on autoregressive RNNs which learns seasonal behaviors and dependencies on given covariates across time series to make probabilistic forecast. By using a shared model across multiple time series (i.e., a multi-task forecast approach), DeepAR can generate forecasts for time series with limited historical data. Temporal Latent Auto-Encoder \cite{nguyen2021temporal} is a recent model designed to tackle high-dimensional multivariate time series forecasting and model cross-series correlations by combining nonlinear factorization of time series and a temporal latent space LSTM forecast model.

 Convolutional neural networks (CNNs) \cite{lecun1998gradient} are a variant of deep neural networks (DNNs) that use a sequence of convolutional layers, pooling layers, and fully-connected layers. CNN-based approaches have been extensively applied and evaluated across a wide spectrum of sequence modeling tasks such as Natural Language Processing (NLP) or speech recognition, and have been shown to outperform RNN-based methods in some cases \cite{johnson2015semi,kalchbrenner2016neural,yin2017comparative}. CNN models have also demonstrated effectiveness in the domain of time series modeling and forecasting \cite{oord2016wavenet,lai2018modeling,borovykh2017conditional,sen2019think}. The temporal convolutional network (TCN) \cite{bai2018empirical} is a CNN-based model using causal convolutions, that is, convolution applied across time such that the convolution output for a given time point depends only on the current and previous time points. The TCN  architecture can take a sequence of any length and map it to an output sequence of the same length via multiple layers of convolution (and padding sequences as necessary).  Notably, TCNs have demonstrated superior empirical performance compared to RNNs and LSTMs in various sequence modeling tasks. For instance, DeepGLO \cite{sen2019think}, a TCN-based model, showed superior results compared to LSTM and DeepAR in time series prediction tasks such as electricity and traffic forecasting. The DeepGLO method is a hybrid model that combines a global matrix
factorization model regularized by a temporal convolution network, along with
another temporal network that can capture local properties of each time-series and
associated covariates. Moreover, the WaveNet architecture has been adapted for conditional time series forecasting \cite{borovykh2017conditional} by using stacked dilated convolutions providing access to a wide range of historical data, and multiple convolutional filters in parallel to separate time series facilitating rapid data processing and exploiting the correlation structure between multivariate time series.
 LSTNet \cite{lai2018modeling} combines a traditional linear autoregressive model with RNN and CNN to extract short-term local dependency patterns among variables and to discover long-term patterns for time series trends. It leverages the strengths of both the convolutional layer to discover the local dependency patterns among multi-dimensional input variables and the recurrent layer to capture complex long-term dependencies.

The recurrence mechanism of RNNs restricts the amount of historical information that can be used for predictions and can complicate training as well, which can limit the capability of such models to capture long-term and complex temporal patterns effectively. Similarly, CNN models typically use dilation to increase their receptive fields, necessitating deeper networks to capture longer-term dependencies. Moreover, these longer-term dependencies are modeled only after several levels of padding and processing in higher layers, which may limit their effectiveness in modeling temporal patterns. To overcome these limitations, Transformer-based models have emerged as a family of models that leverage attention mechanisms instead of recurrence or convolution to model sequential data. Such an approach enables directly modeling the impact of time points within the history window on predicting the future values.  This approach has been proven effective in addressing the flaws of RNNs and CNNs in time series forecasting, especially when large amounts of data are available.

\subsection{Transformer-based time series models}
Transformer-based methods have been proposed to overcome the limitations of previous DNN approaches for time series forecasting and tackle the challenges posed by having modern-day datasets, which are often larger and more complex. The Transformer model \cite{vaswani2017attention} avoids using a recurrence mechanism and relies mainly on a self-attention mechanism to capture cross-sequence (time) interactions and  generate an output. This model achieves very good results in natural language processing, computer vision and time series forecasting \cite{wen2022Transformers,li2019enhancing,wu2020adversarial,lim2021temporal,tang2021probabilistic,lin2021ssdnet,zhou2021informer,wu2021autoformer,liu2021pyraformer,shen2022tcct,cirstea2022triformer,chen2022learning,drouin22tactis,zhou2022fedformer,liu2022non, amazon} which may in part be due to its ability to directly capture long-range dependencies and interactions in sequential data.
Temporal Fusion Transformer (TFT) \cite{lim2021temporal} is a Transformer-based model incorporating variable selection networks to choose pertinent input variables at each time step and static covariate encoders to integrate static features into the model. It processes known and observed inputs using a sequence-to-sequence layer and implements a novel self-attention mechanism with interpretable multi-head attention. This enables TFT to learn long-term relationships across different time steps, making it a powerful tool for capturing complex temporal dependencies in the data. The self-attention mechanism also facilitates the interpretation of feature importance, allowing for the identification of critical factors affecting the forecasting task. Pyraformer \cite{liu2021pyraformer} is a recent Transformer-based model that introduces the pyramidal attention module to describe the temporal dependencies of different ranges leveraging a pyramidal graph and its attention mechanism. Autoformer \cite{wu2021autoformer} replaces the self-attention mechanism in the Transformer by an Auto-Correlation mechanism with dependency discovery and information aggregation at the series level to tackle long-term time series forecasting with intricate temporal patterns. FEDformer \cite{zhou2022fedformer} is a frequency enhanced Transformer, decomposing the input into a trend component with a moving average kernel and a seasonal component. These models are specifically designed for long sequence forecasting. Using the seasonal-trend decomposition structure of FEDformer, DLinear combines it with simple linear layers. Two one-layer linear networks are applied to each component and the two features are summed to get the final prediction. This simple linear approach has   outperformed the state-of-the-art (SOTA) FEDformer for multivariate and univariate forecasting as well as other recent SOTA Transformer-based solutions on different common benchmark datasets.

Note for the multivariate (multiple time series) case these transformer approaches typically treat multiple time series values as part of the input vector for one time point, and an initial embedding linearly maps the combination to a new embedded vector which is subsequently transformed via attention across time points.   Recently PatchTST \cite{Yuqietal-2023-PatchTST} instead applied a shared transformer backbone independently per channel (time series), i.e., in a multi-task manner, and on patches of time points, to transform each channel independently, with a simple linear transformation at the end to get a final prediction for each series, and found this approach to often out-perform the previous channel-mixing transformer forecasting approaches.

\subsubsection{Key model differences with our approach}\label{sec:model_diff}


In summary past transformer-based forecast models were either applied per-series (univariate forecasting) or typically by jointly embedding all variables per time point (multivariate forecasting), with PatchTST using per-series transformation for the Transformer part with a shared network (multi-task approach) but also using a static linear mapping on the transformed variables at the end.   On the one hand, per-series approaches like PatchTST and univariate transformer models often achieve better results than multivariate modeling for many datasets (especially when trained in a multi-task manner across time series) as they can leverage more data (from the multiple time series independently) to train the shared-parameter network and avoid overfitting that can occur with multivariate modeling, especially when there are many time series, and can also be more computationally efficient.  However, on the other hand, per-series models are unable to leverage cross-series information and capture joint time series distribution relationships in forecast outputs.  I.e., while multivariate time series forecast modeling can capture complex relationships between time series it is also  more prone to overfitting and more expensive to train. Furthermore, the static, typically linear, encoding per time point as typically used in multivariate transformer and other neural net forecasting approaches, and as used in PatchTST after independent per-series transformation, limits the relationships that can be captured and modeled between time series - for example, a different kind of transformation, or different information among time series should be shared, depending on the context given by the particular time window being considered.

Instead with our novel transformer forecasting approach, we aim to capture the best of both worlds, of cross-time-series multivariate modeling along with per-series, shared network, multi-task modeling.  Instead of using a static linear transformation of multiple time series values, we apply self-attention across time-series, for the current segment of each time series - to capture the relationships between time series in a dynamic and more flexible and powerful way.  Afterward, a shared transformer network is applied per-series to finish transforming each individual time series and derive the final forecast.  This whole network can then be trained in a multi-task fashion - i.e., applied to each time series individually (each time also feeding in the other time series initially for the cross-series attention transformation part), so it can leverage the benefit of having effectively larger data for training by sharing parameters across time series as in multi-task approaches like PatchTST.
Note that while some past work has explored inter-series attention as well, e.g., for epidemiological analysis \cite{10.1137/1.9781611976700.56}, unlike such prior work, we apply this in a more constrained way as just an initial transformation of directly aligned time series segments only, and critically and uniquely still use a shared multi-task forecast network applied per series after.


\subsubsection{Experiment differences with our approach}


Additionally, most transformer forecasting papers provide results of using these novel transformer methods on widely used, publicly available time series forecasting benchmark datasets, many of which are derived from real-world practical applications, or else on specific domains like epidemiology or logistics. However, the application of these methods to supply chain demand forecasting is very limited. Indeed, most of the papers do not include any retail dataset in the evaluation of their method. The TFT paper \cite{lim2021temporal} is one of the only papers introducing a deep learning method for time series forecasting applied on various domains including a sales forecast dataset in the set of results, but this is not the focus of the study. These models have yet to be explored and demonstrate broad success in supply chain demand  forecasting, which can have challenging characteristics such as sparsity and cross-series effects. As a key distinguishing part of this work we benchmarked many transformer forecast methods, including TFT, with the private dataset provided by the medical device manufacturing company as well as the best of the methods on other retail datasets.

Our method and a diverse set of the models described will be applied on retail datasets to help provide an analysis of the application of these time series methods for supply chain demand forecasting.  This includes an in-depth case study conducted on a supply chain demand forecasting dataset for a medical device manufacturing company. The study provides results for multiple horizons and target metrics, with the objective of catering to the business needs of the company using real-world data. Finally, we evaluated the effectiveness of our proposed approach on two publicly available real-world sales datasets, to provide a better understanding of its performance and its comparison with other methods.

\section{Methodology}
\subsection{Problem Definition}
Let a generic matrix be denoted by a bold capital letter, a vector by a bold lower case letter, and any scalars including individual entries of a vector or matrix by an non-bolded letter. Then, multivariate time-series
 can be represented by a matrix (with each row being an individual time series / variate) and univariate time series by a vector (with each entry being a time point). For a matrix $\Xb$, we denote the i-th column of the matrix with a subscript and lowercase letter $\xb_i$. Let us consider a situation where we have several products for which we want to forecast demand. Each product can be represented as a multivariate time-series denoted by $\Yb^j$, where $j$ represents a specific product. The columns of $\Yb^j$, denoted as $(\yb^j_1, ..., \yb^j_T)$, represent the $T$ time points of our time series, while the rows correspond to the different variates. We consider having $n$ variates. Specifically, at each time point $i$, the vector $\yb^j_i$ consists of the $n$ observed values for the different variables considered. For example, such values consist of useful information such as the sales of the product at a particular location or the price of the product. We consider the problem of forecasting $h$ future values $(y^j_{T+1,l}, ..., y^j_{T+h,l})$ where each $y^j_{i, l}$ represents the value for the dimension $l$ we are interested in, at time point $i$ for product $j$. In this case, we are interested in the same dimension across all future time periods and we call this the label or target variable.  In the demand forecasting case, this may correspond to the sales of the particular product.

\subsection{Preliminaries}
\label{preli}
The Transformer model is based on attention, replacing the recurrent layers most commonly used in encoder-decoder architectures with multi-headed self-attention \cite{vaswani2017attention}.  Through the attention mechanism and the positional encoding, the model is able to appropriately attribute the impact of each element in our input on each other element and compute a representation of a sequence by relating different positions in the sequence. The Transformer architecture consists of stacked self-attention layers and point-wise fully connected layers (i.e., the latter applied per time point, in the case of time series), that sequentially transforms the (vector) representations (also known as encodings), for elements of a sequence with each layer.

\textbf{Attention Mechanism}  \quad The main component of the Transformer responsible for its success is the attention mechanism which can model the relationships between elements in a sequence. The original Transformer paper \cite{vaswani2017attention} uses two different attention mechanisms which are the same at the core: the self-attention mechanism and the encoder-decoder attention. An attention function can be described as mapping a query and a set of key-value pairs to an output, where each query, key and value corresponds to an element in a sequence, e.g., a time point in the case of time series. The input consists of queries and keys of dimension $d_k$, and values of dimension $d_v$ for all elements in a sequence. $\Qb$, $\Kb$ and $\Vb$ are respectively a packed set of queries, a packed set of keys and a packed set of values (with each item in the sequence corresponding to a row). $d_{model}$ is the model dimension to which we project our input. The output for each sequence element is computed as a weighted sum of the values, where the weight assigned to each value is computed by a compatibility function of the query for that element with the corresponding key. In matrix form the set of outputs is given by the following function:
\begin{equation}
    \text{Attention}(\Qb,\Kb,\Vb) = \text{softmax}(\frac{\Qb \Kb^T}{\sqrt{d_k}})\Vb,
\end{equation}
with $\text{softmax}$ applied per row.
In Transformers, these queries, keys, and values are typically determined for each sequence element by linearly projecting the current representation of each element using a learned weight matrix (i.e., which are parameters of the model) for each of the query, key, and value representations, i.e., $\Qb = \Xb \Wb^\Qb$, $\Kb = \Xb \Wb^\Kb$, and $\Vb = \Xb \Wb^\Vb$, where $\Xb$ is the current representation of the sequence.  Thus the output of the self-attention operation results in a new, \textit{transformed} encoding / representation for each element in the sequence.

Instead of performing a single attention function, an extension to this mechanism is to linearly project the queries, keys and values $h$ times with different learned linear projections to $d_k$, $d_k$ and $d_v$ dimensions respectively. The attention function is then performed on each of these projected versions of queries, keys and values, which are concatenated and once again projected. This \textit{multi-head} attention can allow learning different relationships for different parts of a sequence. We label this the MultiHead attention function, described with the following:
\begin{align}
    \text{MultiHead}(\Qb,\Kb,\Vb) &= \text{concat($head_1, ..., head_h$)}\Wb^O
    \\ \text{where } head_i &= \text{Attention}(\Qb\Wb_i^\Qb,\Kb\Wb_i^\Kb,\Vb\Wb_i^\Vb) \nonumber
\end{align}

where $\Wb^\Qb_i \in \mathbb{R}^{d_{model} \times d_k}, \Wb^\Kb_i \in \mathbb{R}^{d_{model} \times d_k}, \Wb^\Vb_i \in \mathbb{R}^{d_{model} \times d_v}$ and $\Wb^O \in \mathbb{R}^{hd_v \times d_{model}}$.  For self attention as used in Transformers, $\Qb = \Kb = \Vb = \Xb$ is the input to $\text{MultiHead}$.

\textbf{Positional Encoding} \quad Positional encoding \cite{DBLP:journals/corr/VaswaniSPUJGKP17, gehring2017convolutional, zerveas2020Transformerbased, anonymous2022mqTransformer} assigns a unique encoding vector to each time step, which is added to the input embedding vector (initial encoding) at that time step. The positional encoding has the same dimension $d_{model}$ as the embeddings, so that the two can be summed. The encoding vector captures information about the position of the time step in the sequence, such as its relative position to other time steps. This enables the model to distinguish between different time steps and understand the temporal ordering of the sequence.

\textbf{Encoder} \quad The encoder is composed of a stack of blocks, each composed of self-attention layers and position-wise feed forward network layers (i.e., the same network applied independently to each sequence element) followed by the residual connection and layer normalization to assist the training stability. The self-attention component of each block of the encoder is in charge of computing the attention weights between all of the elements in the block's input sequence and transforming the elements based on these attentions, as described above.  The encoder thus performs a sequence of transformations to the input sequence representations, and then passes this information onto the decoder for it to roll-out the final predictions.

\textbf{Decoder} \quad The decoder architecture is similar to the encoder architecture. It is also composed of a stack of identical blocks with the same components. In addition to the two sub-layers in each encoder block, the decoder adds a third sub-layer, which performs multi-head attention over the output of the encoder stack as well.  In this way sequential decoder outputs are generated based on both the encoded representations of the input sequence, and the output sequence of the decoder generated so far.

Please refer to \cite{vaswani2017attention} for more details about the Transformer architecture.

\subsection{Model Architecture}
\label{modelarch}
We design a new Transformer-based model, referred to as Inter-Series Transformer, to overcome the different challenging characteristics described in Section \ref{Related Work}.  As mentioned in Section \ref{sec:model_diff}, a key differentiating aspect of our approach is combining controlled, cross-series attention based transformation of different time series, along with per-time-series multi-task modeling via a shared network (for temporal transformation), to capture the best of both worlds.

Note, in the context of supply chain demand forecasting, each time series typically corresponds to a product or product group / category (or product and location combination) and each could also itself be a multivariate time series, incorporating other features like price or promotions.

\begin{figure}[h!]
    \centering
    \includegraphics[scale=.5]{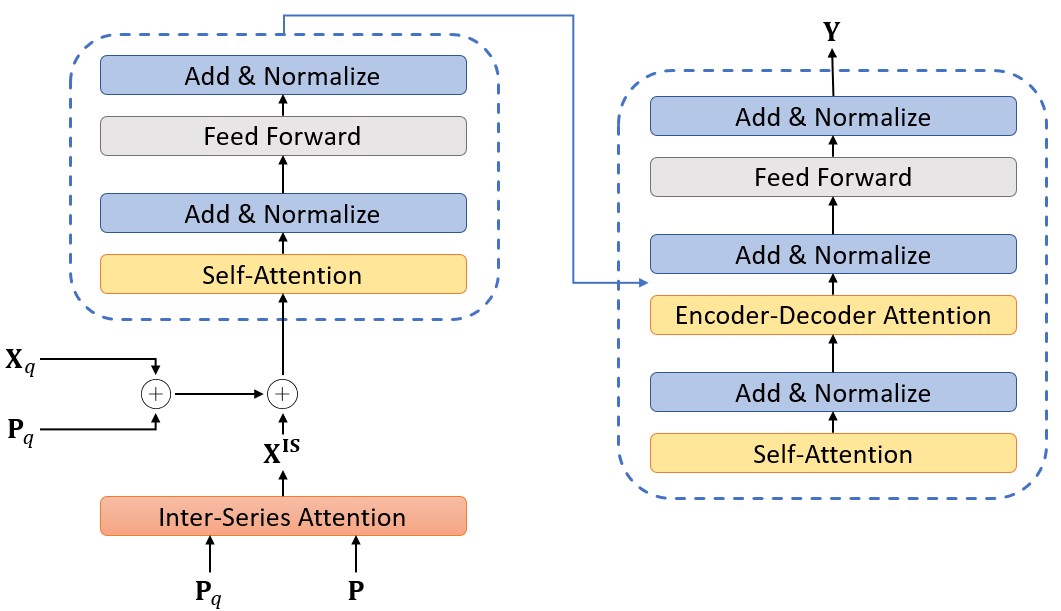}
    \caption{Inter-Series Transformer Diagram with Inter-Series Attention, illustrated here with a single encoder and a single decoder block. Inputs include $\mathbf{P}$ the matrix containing all target time series, $\mathbf{P}_q$ the target time series of product $q$, $\Xb_q$ the feature matrix of product $q$, and $\mathbf{X}^{IS}$ the output from the Inter-Series Attention layer. The circled plus symbol indicates concatenation in the last dimension}
    \label{inter-series layer}
\end{figure}

In our proposed architecture, there are four main differentiating / new components compared to the vanilla Transformer architecture and typical / past Transformer forecasting approaches:
\begin{itemize}
    \item Inter-Series Attention Layer: we introduce a new custom attention layer to get a better informed representation of the target time series by learning dynamics between the different time series / products and incorporating the other time series into the prediction. As shown in Figure \ref{inter-series layer}, this new component is the first layer (or layers) of our custom Transformer and takes all the time series as inputs as detailed in \ref{interserieslayer}.  Note that unlike past Transformer approaches that apply attention just across time, our can leverage attention to capture cross-product / time series effects.  Additionally, unlike applying attention across both all time series and time points at all layers, which could more easily lead to overfitting, ours enables capturing cross time-series effects in a more controlled manner (transforming just the target time series themselves up-front).
    \item Multi-task, shared per-series transformer:  by limiting cross-product / series attention to initial layers and select time series, we control complexity, and enable using a shared Transformer network afterwards to separately transform each individual multivariate time series (e.g., per product), capturing the temporal effects and effectively expanding the amount of data used to train this shared network.
    \item Projection to High Dimensional Representations for various features: we address mixed feature and feature type inputs to create a more comprehensive and informative input for the Transformer by learning separate mappings to high dimensional representations for different feature types.
    \item Abandonment of Positional Encoding: we abandon the positional encoding and we capture relative positioning by defining and incorporating specific features that change with time.
\end{itemize}

\subsubsection{Attention layers}
\label{interserieslayer}
As mentioned in the previous section, the traditional Transformer architecture involves two different attention layers: self-attention and encoder-decoder attention. These two layers allow for the model to learn complex relationships between the different elements in the input/target sequences. However, given that the traditional application was on natural language, this architecture treats each different sequence as somewhat independent to each other. In the context of time-series forecasting within a singular retailer, there are multiple sequences/time series occurring in overlapping time periods, e.g., corresponding to different products or product groups (or product-location combinations). It is important to recognize cross-series effects such as the cannibalization of one product by another. For this reason, we introduce a new attention layer that incorporates other time-series into the prediction for the current time-series.

\textbf{Inter-Series Attention Layer.} \quad We introduce our own custom attention layer which utilizes the same form described above in order to help learn dynamics between different products. We refer to this as Inter-Series attention Layer. As we will see next, this layer takes the Attention function in \ref{preli} and changes the traditional form of the inputs to learn complex dynamics through attention weights between separate time series in addition to different time periods.

The main goal of this layer is to learn attention weights between the context window of the product for which we are providing a forecast and the context windows of all other time series (products) to produce a better informed representation of our desired time series context window. This can also help with sparsity as sparser time series can pay attention to larger-volume time series to improve their predictions, as we will show in our experiment results. To yield a single time series representation from the attention between all products, we must input our target time series context window as the query vector and the context windows of all other time series as the key and value vectors. Therefore, this creates a difference in the input shape between the Inter-Series attention layer and the other attention mechanisms described in \ref{preli}. In the self-attention layer, the query vector and the key/value vectors are all of shape $context\text{-}window \times d_{model}$. In the encoder-decoder attention layer, the query vector and the key/value vectors are respectively of shape $forecast\text{-}window \times d_{model}$ and $context\text{-}window \times d_{model}$. This is intuitive since we are matching like-sequences with each other and relating positions in the sequences to each other based on a high-dimensional representation of our original sequence. In our Inter-Series attention layer, the query vector is instead of shape $1 \times context\text{-}window$, corresponding to the context window of the target
time series, and the key/value vectors stacked are of shape $total\text{-}series \times context\text{-}window$, corresponding to the context windows of all the other time series. We chose to remove the projection to higher dimension since we are only concerned with the target of our other products and want to preserve the notion of a context window in the output. This layer results in an output of dimension $1 \times context\text{-}window$, meant to represent a better-informed version of our target time series within the context window. It is worth noting that the overall architecture can also capture past information through other means such as using other features or incorporating past windows of values.
We augment the original target series in the context window with the informed feature output by the Inter-Series attention Layer and proceed through the model as normal, more formally:
\begin{align}
    \label{eqn1}
    \mathbf{P} &= \text{all products target time series} \\
    \mathbf{P}_{q} &= \text{target time series of product $q$} \nonumber\\
    \Xb_q &= \text{features of product $q$} \nonumber\\
    \mathbf{X}^{IS} &= \text{MultiHead}(\mathbf{P}_{q}, \mathbf{P}, \mathbf{P}) \nonumber\\
    \mathbf{Y} &= Transformer([\mathbf{P}_{q}, \Xb_q, \mathbf{X}^{IS}]) \nonumber
\end{align}
Here, $\mathbf{X}^{IS}$ is the informed feature output of our Inter-Series attention. As detailed in the equations below, the query $\mathbf{P}_{q}$ is the context window of the target time series, and the key/value $\mathbf{P}$ are the context windows of all the time series. As shown in Figure \ref{inter-series layer}, the first layer is our custom Inter-Series attention layer that outputs $\mathbf{X}^{IS}$. This new informed feature $\mathbf{X}^{IS}$ is then concatenated with $\Xb_q$, the matrix feature of product $q$, and $\mathbf{P}_{q}$, its target time series. This concatenation is then fed as input into a Transformer model. If our Inter-Series attention layer is not used, the initial input would simply consist of the concatenation of $\mathbf{P}_{q}$ and $\Xb_q$, resulting in $\mathbf{Y} = Transformer([\mathbf{P}_{q}, \Xb_q])$.

Note, to keep our architecture simple / control complexity, in our experiments we limited our networks to a single layer of this inter-series attention, but it's also possible to apply this for multiple layers to enable a more complex transformation of the target time series based on the other series.

\textbf{Multi-task per-series transformation.} \quad
Note, as depicted in Figure \ref{inter-series layer}, after the inter-series attention layer application, a single transformer network is applied to the combined, transformed representation of the target series, $q$.  This applies time series transformation as is commonly done with Transformers, across time, to capture temporal affects and derive a forecast for future time points.  Notably, this network is shared for all time series (e.g., all products $q$).  That is, the same network parameters are used for all time series, and thus all data can be used to learn the parameters of this shared network - which as mentioned in the Related Work, Section \ref{Related Work}, can often enable better performance / avoid overfitting of multivariate / multi-time-series output approaches.  We call this multi-task application because one shared model is used for each task (product / time series) and applied separately for each, and the model network can also learn to adapt its behavior based on the identifying features and representation elements for each time series, as needed (i.e., as driven by the data).

\subsubsection{Projection to High Dimensional Representation}
 In NLP tasks, inputs are typically sequences of single discrete (categorical) values that need to be quantized through an embedding process to create fixed vector-value representations for each word. Yet, in the context of time series forecasting, features may contain both discrete and real-valued quantities, and of different types (for example multiple different categorical variables), presenting a more complex challenge, not addressed by the original Transformer.  Projecting the features to a meaningful representation is a critical aspect for Transformer models that use real-valued features as well.

To address this issue, we map categorical and real-valued features independently from each other to increased dimensions. For real-valued features, we use a linear layer of weights to learn the optimal mapping. Specifically, we address the issue of low-dimensional real-valued features by replacing the original embedding layer with a linear layer of weights, which allows us to learn the optimal mapping for projecting the inputs to the desired dimensionality. This process creates a set of high-dimensional continuous input features. For categorical features, we employ the original embedding layers to create feature vectors for different features. Finally, the feature vectors and the projected continuous inputs are concatenated before being passed through the Transformer.

Our approach allows us to take advantage of the full range of feature types and dimensions available in the data, creating a more comprehensive and informative input for the Transformer model. By mapping the different types of features independently, we can ensure that each feature is represented optimally in the model, enhancing its ability to capture the complex patterns present in time series forecasting data.

\subsubsection{Abandonment of Positional Encoding}
The positional encoding layer is responsible for assigning a relative position to each element in a given sequence. In the context of time series forecasting, information such as the date and time of day can be critical for producing accurate forecasts. While one approach to incorporating date information into the positional encoding involves adding it as input features and enforcing the positional encoding, this can potentially corrupt the original input and negatively impact the model's stability and ability to learn relationships, as observed in \ref{pos_enco_ablation}. To address this issue, we propose removing the positional encoding altogether and relying solely on the date-time features to capture relative positioning. Specifically,  we map the date into two separate features, one capturing age (year) and the other capturing the month of the year.  Note, for more granular time series, a similar process could be performed for other more granular time segments. To ensure stable training in the deep network, we scale the age feature using the natural $\log$ and the month feature to be between -0.5 and 0.5. This approach allows the model to learn the relative positioning of elements in the sequence based on the known fact that each element follows the prior by one month.

\section{Experimental Setup}

\subsection{Datasets}
\label{datasets}
Our study involves evaluating our proposed model and other time series methods on a private dataset provided by a medical device manufacturing company, which is split into two parts, type 1 and type 2, as explained below. This dataset serves as a guiding case study to improve the limitations of time series forecasting in retail settings.
To further validate our approach, we apply our method to two publicly available retail  datasets, namely Walmart Stores Sales and Walmart M5 \cite{makridakasAccuracy2022}. This allows us to compare the effectiveness of our Inter-Series Transformer model with other time series forecasting techniques, which have been discussed in section \ref{Related Work}, on large-scale sales datasets. By evaluating our approach on both private and public datasets, we can gain a more comprehensive understanding of its potential impact and applicability in real-world retail forecasting scenarios.

\begin{table}[h!]
\centering
\resizebox{\columnwidth}{!}{%
\begin{tabular}{|c|c|c|c|c|}
    \hline
     & Private Dataset Type 1 & Private Dataset Type 2 & Walmart Store Sales & Walmart M5 \\ \hline
    Number of time series & 65 & 50 & 4,410 & 44,280  \\\hline
    Frequency & Monthly & Monthly & Weekly & Daily\\ \hline
    & 1-3 Months & 1-3 Months &  & \\
    Forecast window  & 4-12 Months & 4-12 Months & 39 weeks & 28 days \\
    & 13-24 Months & 13-24 Months &  & \\ \hline
    Metric & wMAPE & wMAPE & RMSE & RMSSE  \\\hline

\end{tabular}%
}
\caption{Information on datasets.}
    \label{tab:datasets}
\end{table}

\textbf{Private Small Retail Dataset.}  \quad  The primary dataset used to evaluate our models is a small retail sales dataset of a medical device manufacturer. Products can be identified with three unique identifiers in a hierarchical structure. Our dataset consists of products across different distribution centers in the world, and each time series corresponds to a specific product at a specific distribution center. We use this approach to avoid having multiple data points at a single time point for a product, which could potentially result in the loss of information through aggregation.

Most of the literature on time series forecasting methods focuses on extremely large datasets, which have been shown to be effective with deep learning techniques that can learn a general understanding of time series mechanisms. However, these approaches may not be applicable in smaller private settings such as ours. Our data consists of two types of products identified at level 1 and resulting in two separate datasets. The first dataset comprises 65 time series corresponding to type 1 products, which exhibit a general increasing trend. The second dataset comprises 50 time series corresponding to type 2 products with a general decreasing trend. Retail datasets often exhibit unpredictable patterns due to external forces that we may not be able to model. Additionally, time series can drop in and out at different time points due to new product introductions, adding to the complexity of forecasting. Our Transformer method employs learned embedding vectors to ensure that new time series can still generate predictions through their identifier features, along with self-attention which is applicable for variable length sequences and multi-task learning so the shared network is trained across the variety of series and this can work effectively on new series.


\textbf{Retail Datasets.}  \quad In addition to conducting experiments on the private dataset provided by a third-party company, we also evaluated the effectiveness of our custom Transformer and other benchmark models on two publicly available Walmart datasets. This allowed us to measure the performance of our model in comparison to other models on much larger datasets, providing insights from our custom attention mechanisms on a massive number of products.

The first dataset we used is the Store Sales forecasting dataset provided by the Walmart Recruiting team for a Kaggle competition. This dataset contains time series of sales for 98 departments at 45 different Walmart stores. While there is no information on specific products, we can consider each department as a separate product, giving us a total of 4,410 time series for potential input during training. One key difference between this dataset and the private one is that we do not have access to any department-specific features, only store-specific ones.

The second dataset we experimented with is the M5 dataset obtained from Walmart \cite{makridakasAccuracy2022}, which is even larger in size compared to the previous Walmart dataset. The data is separated into 3,049 different products sold by Walmart in the US at different stores, and includes aggregated series as well based on the category and department of the product and state location of the stores. This results in a total of 44,280 time series to model on. The significant increase in data makes it difficult to obtain good performance with sufficient epochs, but we present results on a small number of epochs across several algorithms.

\subsection{Training Procedure \& Metrics}

To maximize the performance of the Inter-Series Transformer model, we conducted hyperparameter tuning on several key parameters such as the number of encoder / decoder layers, model dimension, embedding dimension, batch size, and the number of training epochs on validation data. We also experimented with different learning rate scheduling approaches in preliminary study on a subset of the data. Since we did not see much improvement with more exotic approaches, we used a common approach for our experiments. We fixed the learning rate schedule to reduce the learning rate on plateau (by 5\%), with a starting learning rate of 0.0015 and using the Adam optimizer. In the final training setup, we determined the best hyperparameters to use were 2 encoder/decoder layers, 128 model dimension, an embedding dimension of 6 (this is the dimensionality of each time-step in our series), batch size of 64 and 1000 epochs. This resulted in the best validation results for our model. In addition, we similarly fine-tuned the hyperparameters of all neural network methods used for comparison to optimize their performance, ensuring consistency and soundness of the results presented in Section \ref{results}. The inter-series transformer required its own tuning-framework as prototype code was developed, but since the others are popular models we were able to use AutoML directly to train several versions in a parallelized fashion and compare. The shown results are the best results after hyperparameter tuning.

As shown in Table \ref{tab:datasets}, the evaluation window depends on the dataset. For the private dataset, we evaluated our model on three distinct time ranges: short-range (1-3 months), mid-range (4-12 months), and long-range (13-24 months). For the Walmart Store Sales, the forecast window is 39 weeks, while for Walmart M5, it is 28 days. To obtain the final score for each range, we calculated the average of scores / metrics across all evaluation frequencies within the respective period.

The choice of metric also depends on our datasets as well as the recommendations given to us by the private retailer providing the medical dataset - i.e., the metrics the retailer wanted to use for evaluation and uses internally. Precise definitions of the metrics used are given below. Regarding the metrics used in the Kaggle competitions, while the Walmart Store Sales competition employed weighted-MAE, we use RMSE instead as that competition was an older one, and future modern competitions have preferred squared-error metrics, such as the M5 competition.  Furthermore, weighting the absolute errors ends up doing something similar to MSE anyway which naturally weights larger errors more and on average larger magnitude series.
 Additionally as the data is at an aggregate level scaling is likely not necessary, and thus RMSE should work comparably to the original competition metric.
 For the Walmart M5 competition, we employ the Root Mean Squared Scaled Error (RMSSE) as used in the competition - where scaling is applied to account for the greater variation and sporadic nature and sparsity in the series, enabling directly comparing scores across series.  However, unlike the competition we do not further weight each series by recent training sales volumes.  The main reason is this weighting biases the scoring to more heavily weight the fewer, high level aggregate series, which become smoother and easier to predict as the aggregation increases. This can be seen from the final weighted average scores in the competition, as well as the average high level aggregate scores, having significantly lower RMSSE than the granular, low-level series average scores \cite{makridakasAccuracy2022}.
 This evaluation would defeat the purpose of applying and evaluating our method for which we want to see if it can improve the performance across diverse retail series and in particular on the sparse, sporadic series in the lower levels of the hierarchy making up the majority of the time series.  Therefore we use the unweighted RMSSE to obtain a fair comparison of overall predictive performance across the diverse set of series.



\begin{equation}
    \text{wMAPE} = \frac{\sum_{\text{all series and timesteps}}\lvert\text{Actual} - \text{Forecast}\rvert}{\sum_{\text{all series and timesteps}}\lvert\text{Actual}\rvert}
\end{equation}

\begin{equation}
    \text{RMSE} = \sqrt{\frac{\sum_{\text{all series and timesteps}}(\text{Actual} - \text{Forecast})^2}{mT}}
\end{equation}
Here, $m$ is the number of series and $T$ the number of timesteps.

\begin{equation}
    \text{RMSSE} = \sqrt{\frac{1}{h}\frac{\sum^{n+h}_{t=n+1}(y_t - \hat{y}_t)^2}{\frac{1}{n-1}\sum_{t=2}^n(y_t-y_{t-1})^2}}
\end{equation}
Here RMMSE is shown for one series, $y_t$ is the actual future value at time t, $\hat{y_t}$ is the predicted value, n is the number of historical observations and h is the horizon.

\section{Results}
\label{results}
We applied our custom Inter-Series Transformer model, along with traditional time series methods and neural network models, to the datasets described in \ref{datasets}. In this section, we present and compare the results of our approach with those of benchmark models.

\begin{table}[h!]
\centering
\resizebox{\columnwidth}{!}{%
\begin{tabular}{  | cc | ccc | ccc | c|}
        \hline
         \multicolumn{2}{|c|}{Method} & \multicolumn{3}{c|}{Type 1}
         & \multicolumn{3}{c|}{Type 2}
         & \multicolumn{1}{c|}{Training Time} \\
         && 1-3 Months & 4-12 Months & 13-24 Months & 1-3 Months & 4-12 Months & 13-24 Months & in seconds \\\hline
        \multicolumn{2}{|c|}{Baseline} & 21.7\% & 49.3\% & 71.3\% & 16.8\% & 27.8\% & 58.2\% & N/A   \\\hline

  & \multicolumn{1}{|c|}{SES}  & 65.3\% & 73.2\% & 121.1\% & 55.8\% & 63.8\% & 109.5\%  & 10s \\\cline{2-9}
  Traditional time series & \multicolumn{1}{|c|}{HW}  & 60.2\% & 65.8\% & 96.6\% & 53.5\% & 61.8\% & 93.0\% & 27s \\\cline{2-9}
  & \multicolumn{1}{|c|}{ARIMA}  & 71.9\% & 77.0\% & 109.3\% & 56.9\% & 68.5\% & 132.7\%  & 478s\\\hline
  &\multicolumn{1}{|c|}{Feed-Forward} & 68.2\% & 69.0\% & 109.1\% & 54.8\% & 68.3\% & 105.2\% &   448s \\\cline{2-9}
  &\multicolumn{1}{|c|}{DeepAR}  & 37.6\% & 39.3\% & 37.8\% & 21.9\% & 23.5\% & 35.6\% & 897s \\\cline{2-9}
  Neural Networks & \multicolumn{1}{|c|}{GluonTS Transformer} & 35.0\% & 36.6\% & 38.4\% & 17.0\% & 19.0\% & \textbf{19.1\%} & 3537s\\\cline{2-9}
  Models & \multicolumn{1}{|c|}{TFT}  & 30.0\% & 28.6\% & 31.4\% & 17.0\% & 20.0\% & 21.2\% & 3698s \\\cline{2-9}
  & \multicolumn{1}{|c|}{FEDFormer}  & 68.2\% & 72.2\% & 85.7\% & 61.5\% & 70.9\% & 80.2\% & 2843s \\\cline{2-9}
  & \multicolumn{1}{|c|}{DLinear}  & 31.0\% & 48.3\% & 64.8\% & 26.6\% & 37.1\% & 50.0\% & 1981s \\\cline{2-9}
   & \multicolumn{1}{|c|}{PatchTST}  & 16.5\% & 26.2\% & \textbf{29.8\%} & 14.1\% & \textbf{18.1}\% & 35.8\% & 3291s \\\hline
    \multicolumn{2}{|c|}{Inter-Series Transformer} & \textbf{14.7\%} & \textbf{24.9\%} & 59.2\% & \textbf{12.9\%} & 18.8\% & 40.3\% & 4137s   \\\hline
\end{tabular}%
}
\caption{wMAPE Results of Methods discussed on Type 1 and Type 2 Medical Device Datasets Separated by Forecast Range.}
    \label{tab:t1}
\end{table}

 Regarding our private medical device manufacturing dataset, we also include the results of a baseline method, which is the forecast provided to us by a third-party. It was obtained using a business-oriented approach, which was derived from several stakeholder meetings, as opposed to our data-driven approach. As shown in Table \ref{tab:t1}, the Inter-Series Transformer outperforms the baseline and traditional time series methods across all forecast ranges, as well as neural networks models for short-term forecasting (1-3 Months) and mid-term forecasting (4-12 Months).

We first analyze the results from some of the more traditional time series forecasting models. The methods used include Holt-Winters, Autoregressive Integrated Moving Average (ARIMA), and Simple Exponential Smoothing (SES) models. For all of these examples, we fit individual models to each time series in our private dataset, where a time series pertains to the combination of a specific product at a specific distribution center. Comparing the results of these three models to the baseline, it is clear that the traditional methods alone are not sufficient to reach the desired levels of accuracy on either type of device.

We then experiment with more complex methods that rely on neural networks. The feed-forward method performs worse than some of the traditional methods since there is no form of sequential modeling using this technique. We also explore models discussed in Section \ref{Related Work}, including DeepAR, GluonTS Transformer, and TFT. GluonTS Transformer is representative of the base Transformer approach to forecasting and serves as a starting Transformer model for comparison.  It has the same base architecture as our Inter-Series Transformer, of Transformer encoder-decoder, without all the enhancements / modifications we introduced, and is fed the same features and inputs. While these methods do not outperform the Inter-Series Transformer for short-range and mid-range forecasting, they perform better for long-term forecasting (13-24 months), with TFT achieving the best overall results in this range. In our approach, we prioritize capturing cross-series effects and addressing sparsity in the retail setup.  The longer-term targets are set for longer aggregate periods and further into the future. Consequently, there may be less advantage in adapting long-term predictions solely based on the recent behavior of other series. That is, recent cross time series history may be more beneficial for nearer-term forecasting, at less-aggregated horizons, and different time series may be more likely to diverge further into the future, or at least not follow short-term patterns.  As such the Inter-Series modeling we introduce may add additional complexity that hurts overall accuracy in these long-term aggregate predictions. On the other hand, one of TFT's notable contributions was the introduction of a multi-head attention mechanism with an additive aggregation of the different heads, specifically designed to capture long-term dependencies, which could explain its superior performance in long-term predictions.

Additionally, we initially applied two of the most recent, state-of-the-art, time series forecasting models to our datasets, FEDformer, and DLinear, and compared their performances with the other methods. Although these models are designed to improve performance for long-term time series forecasting, the FEDformer model does not outperform the baseline for any range level, and the DLinear model outperforms it by a small margin for long-term forecasting. However, the proposed Inter-Series Transformer achieves better results across all ranges.
This may be in part due to designing and testing these approaches on non-retail data that has different and often more predictable temporal patterns as opposed to retail / count data type of time series.  This further illustrates the importance of testing algorithms on diverse use cases and evaluating them with targeted case studies for these different domains such as we provide here for retail data.

Subsequent to our initial experiments and write up, PatchTST was released as a new state-of-the-art transformer forecasting method, particularly for longer time ranges, so we compared results with this method as well.
For Type 1, PatchTST does perform better at the long time range but is out-performed by our method in the shorter 2 ranges, and for Type 2 our method out-performs it at the shortest range, has comparable performance at the medium range, and is out-performed by it at the longer range.  This demonstrates PatchTST can be more effective at longer ranges, and similar explanation and conclusion as given with TFT above can be made.

\begin{table}[h!]
\centering
\begin{tabular}{ |c|c|c| }
        \hline
         Method
         & Walmart Store Sales
         & Walmart M5  \\
         & RMSE  & RMSSE  \\

  \hline
  DeepAR & 0.651 & 1.010 \\
  \hline
  TFT & 0.528 & 1.083 \\
  \hline
  DLinear & 0.556 & 0.956 \\
  \hline
  HW & 1.190 & 1.581 \\
  \hline
  Inter-Series Transformer & 0.511 & \textbf{0.809} \\
  \hline
  PatchTST & \textbf{0.492} & 0.976 \\
  \hline
\end{tabular}
\caption{Results of Methods discussed on Walmart Retail Datasets.}
    \label{tab:t2}
\end{table}

Furthermore, we evaluated the performance of our Inter-Series Transformer on the two Walmart datasets described in Section \ref{datasets}.
The Walmart M5 dataset is very large and at a more granular level, while the Walmart Store Sales dataset is smaller and at a more aggregate level, though both are larger than our private retail dataset, so collectively these set of datasets make up a suitable collection to evaluate and analyze our method across a range of different retail forecasting use cases. I.e., these two datasets allow us to measure the effectiveness of our approach and compare it to other time series models on a more diverse set of retail data.

As seen in Table \ref{tab:t2}, we compare the performance of our model with that of one traditional time series method, Holt-Winters, as well as deep learning models: DeepAR, TFT, and DLinear, i.e., the best-performing model of each category previously, as well as PatchTST which performed well in some cases on the previous, target dataset, and is the most recent state-of-the-art comparison method. Based on these experiments, the Inter-Series Transformer outperforms all of these methods on the more granular M5 Walmart dataset, and is second place and close to the best (PatchTST) on the more aggregate and simpler Walmart Store Sales dataset (which also had less features available to make use of). Therefore, the Inter-Series Transformer is a promising approach that can achieve high performance on datasets of various sizes, as compared to the benchmarks.  For the case of the Walmart Store Sales dataset, the lack of features and similarity and regularity of the different series due to aggregate (department) level forecasting, along with the relatively longer forecast window, can explain why the unique components of our approach were not able to lead to improved performance over the long-term, basic per-series forecast approach of PatchTST.  Nevertheless our method was still very competitive even on this dataset and still obtained the second best score.

\begin{figure}[h!]
    \centering
    \includegraphics[scale=.468]{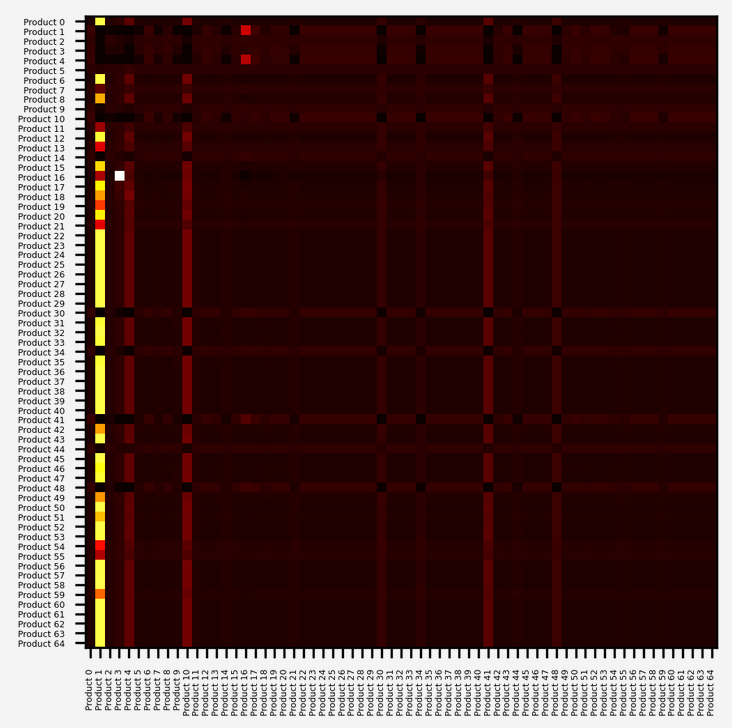}
    \caption{Attention weights learned between products / time series of type 1 - for one prediction.  Each row shows the attention weights for each other series across the columns, for that target series.   Lighter color indicates a higher value.}
    \label{attw}
\end{figure}

Finally, Figure \ref{attw} highlights how the time series of the different products of type 1 are learning to pay attention among themselves - illustrating cross-series attention weights for one prediction / time point, and provides an example of interpretability achievable with this approach.

\begin{figure}[h!]
    \centering
    \includegraphics[scale=.27]{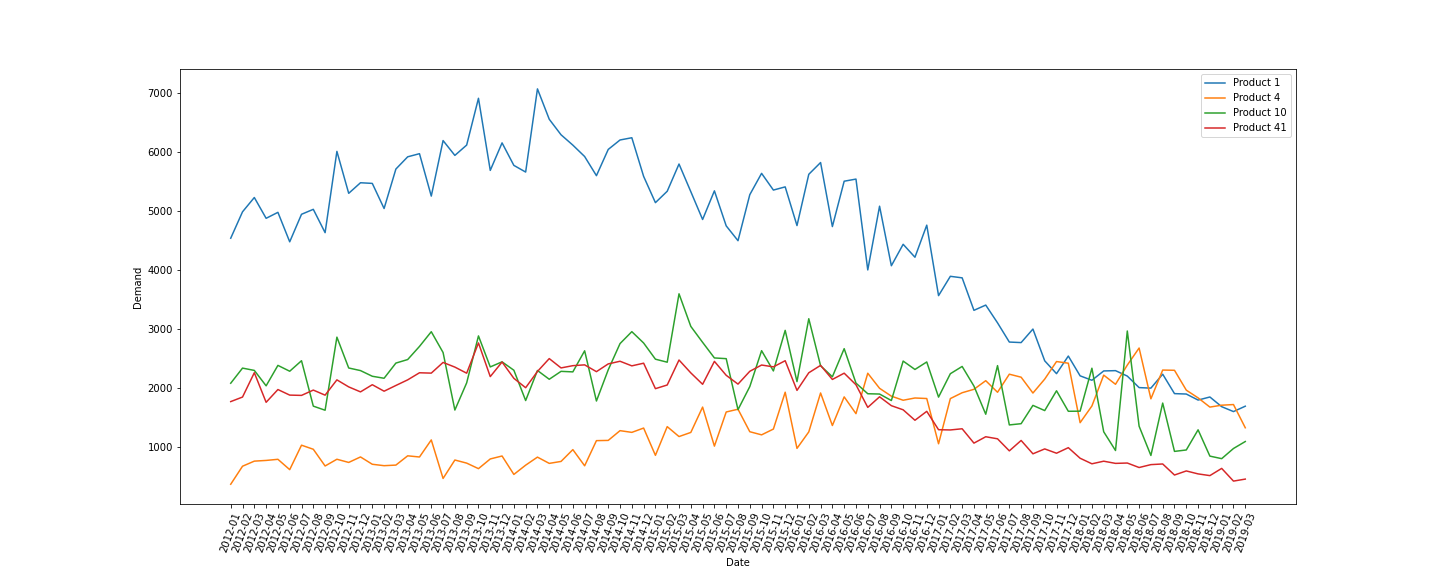}
    \caption{Example of high-volume products. Products 1, 4, 10 and 41 have most of the attention weight from sparser time series as shown in Figure \ref{attw}.}
    \label{highvolts}
\end{figure}

\begin{figure}[h!]
    \centering
    \includegraphics[scale=.27]{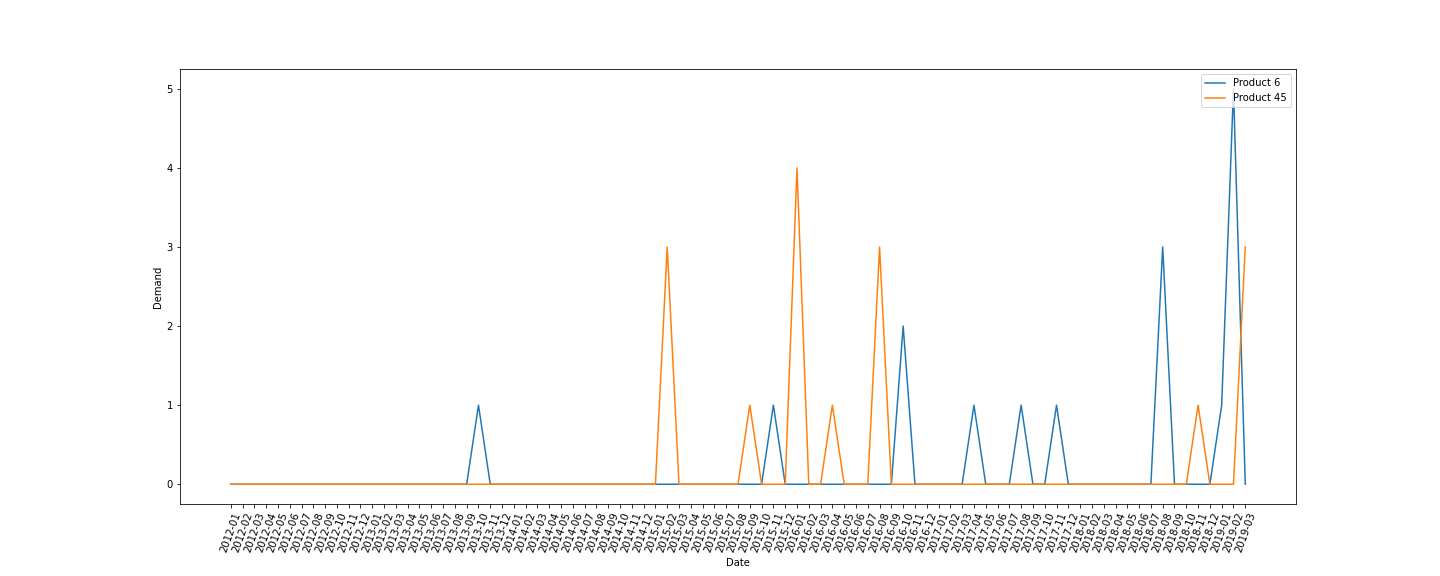}
    \caption{Example of low-volume and sparse products. Many time series, such as products 6 and 45, depend highly on products 1, 4, 10 and 41 as shown in Figure \ref{attw}.}
    \label{lowvolts}
\end{figure}

\begin{figure}[h!]
    \centering
    \begin{subfigure}{0.48\textwidth}
        \centering
        \includegraphics[scale=.5]{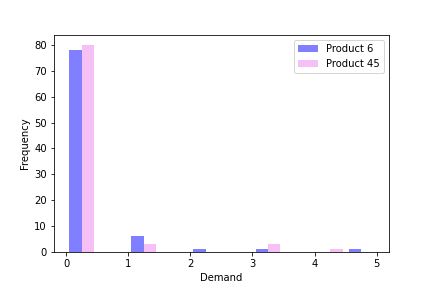}
        \caption{Distribution of products 6 and 45.}
        \label{histlowvol}
    \end{subfigure}
    \hfill
    \begin{subfigure}{0.48\textwidth}
        \centering
        \includegraphics[scale=.5]{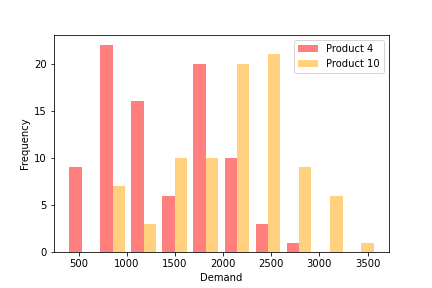}
        \caption{Distribution of products 4 and 10.}
        \label{histhighvol}
    \end{subfigure}
    \caption{Comparison of value distribution for sparse (products 6 and 45) and high-volume (products 4 and 10) time series.}
    \label{dist}
\end{figure}

By analyzing the corresponding products and time series, we observe that time series with higher attention weights generally have greater volume and stability. In contrast, sparser time series, characterized by lower volume and fewer observations, rely more on these high-volume time series. Figure \ref{attw} illustrates how a few series, specifically products 1, 4, 10 and 41, attract most of the attention weight from other series. These series are associated with high volumes, as seen in Figure \ref{highvolts}. Conversely, the other products that focus their attention on these high-volume products, like products 6 and 45, exhibit sparser series, as illustrated in Figure \ref{lowvolts}, and demonstrate a skewed-value distribution, as emphasized by Figure \ref{dist}. These observations support our hypothesis that introducing the Inter-Series Attention layer can help address issues of sparsity and skewed-value distribution by allowing products to learn from larger volume time series, thereby improving predictions.

\section{Analysis \& Ablation Study}

We are able to improve on the existing benchmark for the private dataset provided to us by incorporating the new components described in \ref{modelarch}. Yet, the effectiveness of our model could be limited to the specific evaluation period used, so we apply a time series style cross-validation technique \cite{Cerqueira_2020} to evaluate its robustness. Specifically, for each evaluation period, we trained individual models prior to that period and evaluated them on the evaluation period, and calculated the average of the results across the periods. The training data included all historical data prior to the evaluation period, and we treated the separate evaluation periods as test sets. Although our training datasets were not all of the same sizes and some contained older data than others, we attempted to balance them by limiting how far back in time the model could begin training.

\begin{figure}[h!]
    \centering
    \includegraphics[scale=.6]{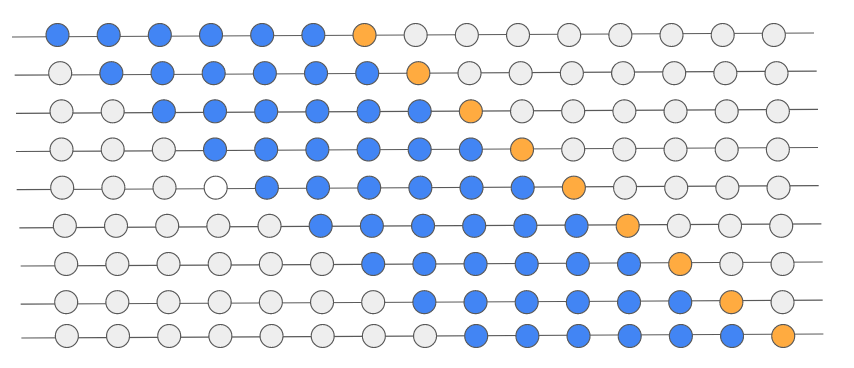}
    \caption{Sliding evaluation framework for the case of 1-Month forecast.}
    \label{fig:robustness}
\end{figure}

To check the robustness of our method, we measure performance across time-periods without tuning any parameters and shift our evaluation set. We provide results across several evaluation periods. Table \ref{tab:rob} shows that our custom approach can successfully perform across different time periods and is not simply overfitting or performing well for a particular period. The results were calculated using the sliding window method illustrated in Figure \ref{fig:robustness}. These results indicate that the model can significantly outperform the baseline forecasts across different time periods.

\begin{table}[h!]
    \centering
    \resizebox{\columnwidth}{!}{
        \begin{tabular}{|c|c|c|c|c|c|}
            \hline
            Forecast Start Date & Forecast Range & Metric & Baseline Result & Inter-Series Transformer & \% Improvement \\
            \hline
            06/2017 & 1-3 Months & wMAPE & 20.35\% & \textbf{12.03\%} & 8.32\% \\

            \hline
            03/2018 & 1-3 Months & wMAPE & 46.54\% & \textbf{21.11\%} & 25.43\% \\

            \hline
            09/2018 & 1-3 Months & wMAPE & 55.21\% & \textbf{35.90\%} & 19.31\% \\

            \hline

        \end{tabular}
    }
    \caption{Inter-Series Transformer Results Compared to Baseline Results over several evaluation periods for short-range forecast.}
    \label{tab:rob}
\end{table}

Furthermore, we conducted an ablation study focusing on two key aspects of the Transformer model:
\begin{itemize}
\item We explored the different options of projecting our features to a higher dimension to optimize learning.
\item We investigated the effectiveness of not using positional encoding, as well as alternative approaches to capture temporal ordering in a sequence.
\end{itemize}
The results presented in this section pertain to the private dataset. In addition, we introduce a new metric, $wBias$, which was added to meet specific needs of the medical device manufacturing company in the longer horizons.
\begin{equation}
    \text{wBias} = \frac{\sum_{\text{all series}}(\text{series volume} \times \lvert\text{mean(Actual)-mean(Forecast)}\rvert)}{\sum_{\text{all series}}}
\end{equation}

\subsection{Projection to High Dimensional Representation}
In the context of time series forecasting, we are not dealing entirely with discretized quantities anymore and have features that contain a real quantifiable value before any form of embedding. We are also not exclusively dealing with one type of feature (e.g., discrete words) but have access to a variety of features of different types, including categorical features such as identifier features and continuous features as historical price. These differences leave us with many options for how we choose to handle projecting our features to a higher dimension for the deep network.

For continuous real-valued features, we addressed the issue of our low-dimension by replacing the input embedding layer with a linear layer. This allows us to learn the optimal way of mapping our input to our desired dimensionality.
 While we treat the date of our input as two separate continuous features, we still have an important categorical feature that we must address: the identifier. Since we have a hybrid identifier, we experiment with two different approaches to embedding this information. The first approach was to use the combined identifier as a single feature and employ a single embedding layer to map a time series to a feature vector. The second approach was to separate the two identifier features and map them with the help of two separate embedding layers to create two feature vectors. The second option offers the possibility to learn similarities between the same product at different centers or vice-versa. These feature vectors are created alongside but independently from the projected input, which we replaced the original embedding layer with. The projected continuous inputs and the categorical feature vectors are then concatenated before being passed through the Transformer.

We also experimented with different ways of incorporating the categorical ID features into the input. This required modifying the way we projected our input to a higher dimension in order to optimize learning. Ultimately, the results in Table \ref{tab:highdim} showed that embedding the two separate ID features (location, product ID) independently yielded the best results since the model can optimize both embeddings separately.

\begin{table}[h!]
    \centering
    \resizebox{\columnwidth}{!}{
        \begin{tabular}{|c|c|c|c|}
             \hline
             Metric &  Joint Embedding &  Separate Embedding with one ID Feature & Separate Embedding with two ID Feature\\
             \hline
             1-3 Months wAvgMAPE & 25.94\% & 24.64\% &  \textbf{24.58\%}\\
             1-3 Months wMAPE & 25.68\% & 23.86\% & \textbf{23.72\%}\\
             \hline
             4-12 Months wBias & 1632.75 & 1627.16  & \textbf{1377.52}\\
             4-12 Months wMAPE & 50.78\% & 53.56\% & \textbf{49.57\%} \\
             \hline
             13-24 Months wBias & 1038.61 & 813.806 & \textbf{788.290} \\
             13-24 Months wMAPE & 58.12\% & 66.45\% & \textbf{60.38\%} \\
             \hline
        \end{tabular}
    }
    \caption{High-Dimensional Projection Experiments.}
    \label{tab:highdim}
\end{table}

\subsection{Positional Encoding}
\label{pos_enco_ablation}
The positional encoding aspect of Transformers is critical in NLP applications where there is no other notion of a time point, but in time series applications, we have the benefit of having actual time features. Because of this, we experiment with using the positional encoding alongside our time information and simply excluding the positional encoding.

Regarding the positional encoding, one attempt to strengthen it was to simply provide any important date information as features of our input as well as enforce the positional encoding. While this resulted in more accurate forecasts compared to benchmark models, the results were still not satisfactory. The positional encoding vector which is generated is added to our original sequence to get the input which will continue through our model. The fact that we alter our initial input with this positional encoding could impact the stability of our training and ability to learn relationships effectively. The second approach is to remove the positional encoding altogether to combat the fact that it is corrupting our original input. The model should be able to learn the relative positioning of elements just as easily through the date feature provided, particularly since the elements that follow each other in a sequence will always be one month after the prior. In order to facilitate the learning of this fact, we decided to map the date into two separate features: one which captures the age (or year) and one which captures the month of the year.

We conducted experiments to compare two possibilities: first, adding positional encoding to the categorical time features, and second, mapping the date into two continuous time features while removing positional encoding. As shown in Table \ref{tab:posencoding}, we found that utilizing continuous time features without positional encoding yielded the best results. It appeared that positional encoding mostly did not provide any additional information and only added complexity to the training process for the model. Therefore, we made the decision to exclude positional encoding and instead rely on continuous features.

\begin{table}[h]
    \centering
    \resizebox{\columnwidth}{!}{
    \begin{tabular}{|c|c|c|}
         \hline
         Metric &  Categorical Time Feature with Positional Encoding & Continuous Time Features without Positional Encoding \\
         \hline
         1-3 Months wAvgMAPE & 30.1\% & \textbf{25.94\%} \\
         1-3 Months wMAPE & 31.0\% & \textbf{25.68\%} \\
         \hline
         4-12 Months wBias & \textbf{1401.75} & 1632.75 \\
         4-12 Months wMAPE & 53.7\% & \textbf{50.78\%}\\
         \hline
         13-24 Months wBias & \textbf{596.910} & 1038.61 \\
         13-24 Months wMAPE & 59.3\% & \textbf{58.12\%} \\
         \hline
    \end{tabular}
    }
    \caption{Positional Encoding Experiments.}
    \label{tab:posencoding}
\end{table}

\section{Conclusion}
Although Transformer-based models have shown remarkable performance in time series forecasting, their application to supply chain demand forecasting is still limited. Therefore, we introduce the Inter-Series Transformer, a novel architecture that incorporates a new attention layer, the Inter-Series attention layer, to capture cross-series effects in a controlled manner  while still enabling multi-task Transformer network application to generate forecasts per-series, and address the sparsity and overfitting challenges in supply chain-related time series data. We evaluate the proposed model on a private dataset from a medical device manufacturer as well as larger retail datasets, and the results demonstrate that the Inter-Series Transformer outperforms traditional time series models and neural network models, and is competitive with and often out-performs a number of prior state-of-the-art neural network and Transformer forecast models in most cases. Furthermore, the model's robustness is evidenced by its ability to outperform the baseline for various start evaluation times. We also found that positional encoding was not necessary for higher performance and instead recommend adding date features. Our proposed model presents a promising approach for supply chain demand forecasting, with implications for enhancing supply chain management and operations.

Several avenues for future work can build on the promising results of the Inter-Series Transformer in supply chain demand forecasting. One potential direction could be to explore extending the Inter-Series Transformer's attention mechanism to features of the other time series data in addition to the target time series. This would allow for a more comprehensive analysis of the relationships between different features and potentially lead to more accurate forecasting results. Additionally, an interesting direction for further research would be to investigate the potential benefits of increasing the depth of the Inter-Series Transformer by adding multiple Inter-Series attention layers. This could allow to capture more complex nonlinear transformations of inputs, potentially leading to even better forecasting performance. These directions present exciting opportunities to further improve and expand upon the proposed Inter-Series Transformer architecture for time series forecasting.


\bibliography{mybibfile.bib}

\newcommand{\noopsort}[1]{} \newcommand{\printfirst}[2]{#1}
  \newcommand{\singleletter}[1]{#1} \newcommand{\switchargs}[2]{#2#1}
\begin{thebibliography}{10}
\expandafter\ifx\csname url\endcsname\relax
  \def\url#1{\texttt{#1}}\fi
\expandafter\ifx\csname urlprefix\endcsname\relax\def\urlprefix{URL }\fi
\expandafter\ifx\csname href\endcsname\relax
  \def\href#1#2{#2} \def\path#1{#1}\fi

\bibitem{10.2307/20202189}
R.~Fildes, K.~Nikolopoulos, S.~F. Crone, A.~A. Syntetos,
  \href{http://www.jstor.org/stable/20202189}{Forecasting and operational
  research: A review}, The Journal of the Operational Research Society 59~(9)
  (2008) 1150--1172.
\newline\urlprefix\url{http://www.jstor.org/stable/20202189}

\bibitem{KIM2003307}
K.~jae Kim,
  \href{https://www.sciencedirect.com/science/article/pii/S0925231203003722}{Financial
  time series forecasting using support vector machines}, Neurocomputing 55~(1)
  (2003) 307--319, support Vector Machines.
\newblock \href
  {http://dx.doi.org/https://doi.org/10.1016/S0925-2312(03)00372-2}
  {\path{doi:https://doi.org/10.1016/S0925-2312(03)00372-2}}.
\newline\urlprefix\url{https://www.sciencedirect.com/science/article/pii/S0925231203003722}

\bibitem{chatfield2000time}
C.~Chatfield, \href{https://books.google.com/books?id=PFHMBQAAQBAJ}{Time-Series
  Forecasting}, CRC Press, 2000.
\newline\urlprefix\url{https://books.google.com/books?id=PFHMBQAAQBAJ}

\bibitem{mitchell1980need}
T.~M. Mitchell, The need for biases in learning generalizations, Tech. Rep.
  CBM-TR-117, Rutgers University (1980).

\bibitem{gordon1995evaluation}
D.~F. Gordon, M.~Desjardins, Evaluation and selection of biases in machine
  learning, Machine learning 20~(1) (1995) 5--22.

\bibitem{mckenzie1984general}
E.~McKenzie, General exponential smoothing and the equivalent arma process,
  Journal of Forecasting 3~(3) (1984) 333--344.

\bibitem{hyndman2008forecasting}
R.~Hyndman, A.~B. Koehler, J.~K. Ord, R.~D. Snyder, Forecasting with
  exponential smoothing: the state space approach, Springer Science \& Business
  Media, 2008.

\bibitem{box1968some}
G.~E. Box, G.~M. Jenkins, Some recent advances in forecasting and control,
  Journal of the Royal Statistical Society. Series C (Applied Statistics)
  17~(2) (1968) 91--109.

\bibitem{hyndman2018forecasting}
R.~J. Hyndman, G.~Athanasopoulos, Forecasting: principles and practice, OTexts,
  2018.

\bibitem{rumelhart1986learning}
D.~E. Rumelhart, G.~E. Hinton, R.~J. Williams, Learning representations by
  back-propagating errors, nature 323~(6088) (1986) 533--536.

\bibitem{lipton2015critical}
Z.~C. Lipton, J.~Berkowitz, C.~Elkan, A critical review of recurrent neural
  networks for sequence learning, arXiv preprint arXiv:1506.00019.

\bibitem{hochreiter1997long}
S.~Hochreiter, J.~Schmidhuber, Long short-term memory, Neural computation 9~(8)
  (1997) 1735--1780.

\bibitem{gers2001applying}
F.~Gers, D.~Eck, J.~Schmidhuber, Applying lstm to time series predictable
  through time-window approaches, in: Proceedings of the International
  Conference on Artificial Neural Networks, 2001, pp. 669--676.

\bibitem{salinas2020deepar}
D.~Salinas, V.~Flunkert, J.~Gasthaus, T.~Januschowski, Deepar: Probabilistic
  forecasting with autoregressive recurrent networks, International Journal of
  Forecasting 36~(3) (2020) 1181--1191.

\bibitem{bandara2020}
K.~{Bandara}, C.~{Bergmeir}, H.~{Hewamalage}, Lstm-msnet: Leveraging forecasts
  on sets of related time series with multiple seasonal patterns, IEEE
  Transactions on Neural Networks and Learning Systems (2020) 1--14.

\bibitem{lai2018modeling}
G.~Lai, W.-C. Chang, Y.~Yang, H.~Liu, Modeling long-and short-term temporal
  patterns with deep neural networks, in: The 41st international ACM SIGIR
  conference on research \& development in information retrieval, 2018, pp.
  95--104.

\bibitem{salinas2019high}
D.~Salinas, M.~Bohlke-Schneider, L.~Callot, R.~Medico, J.~Gasthaus,
  High-dimensional multivariate forecasting with low-rank gaussian copula
  processes, in: Proceedings of the 33rd International Conference on Neural
  Information Processing Systems, 2019, pp. 6827--6837.

\bibitem{smyl2020hybrid}
S.~Smyl, A hybrid method of exponential smoothing and recurrent neural networks
  for time series forecasting, International Journal of Forecasting 36~(1)
  (2020) 75--85.

\bibitem{nguyen2021temporal}
N.~Nguyen, B.~Quanz, Temporal latent auto-encoder: A method for probabilistic
  multivariate time series forecasting, in: Proceedings of the AAAI Conference
  on Artificial Intelligence, Vol.~35, 2021, pp. 9117--9125.

\bibitem{vaswani2017attention}
A.~Vaswani, N.~Shazeer, N.~Parmar, J.~Uszkoreit, L.~Jones, A.~N. Gomez,
  {\L}.~Kaiser, I.~Polosukhin,
  \href{https://proceedings.neurips.cc/paper/2017/file/3f5ee243547dee91fbd053c1c4a845aa-Paper.pdf}{Attention
  is all you need}, in: I.~Guyon, U.~V. Luxburg, S.~Bengio, H.~Wallach,
  R.~Fergus, S.~Vishwanathan, R.~Garnett (Eds.), Advances in Neural Information
  Processing Systems, Vol.~30, Curran Associates, Inc., 2017.
\newline\urlprefix\url{https://proceedings.neurips.cc/paper/2017/file/3f5ee243547dee91fbd053c1c4a845aa-Paper.pdf}

\bibitem{lin2022survey}
T.~Lin, Y.~Wang, X.~Liu, X.~Qiu,
  \href{https://www.sciencedirect.com/science/article/pii/S2666651022000146}{A
  survey of transformers}, AI Open\href
  {http://dx.doi.org/https://doi.org/10.1016/j.aiopen.2022.10.001}
  {\path{doi:https://doi.org/10.1016/j.aiopen.2022.10.001}}.
\newline\urlprefix\url{https://www.sciencedirect.com/science/article/pii/S2666651022000146}

\bibitem{wen2022Transformers}
Q.~Wen, T.~Zhou, C.~Zhang, W.~Chen, Z.~Ma, J.~Yan, L.~Sun, Transformers in time
  series: A survey, arXiv preprint arXiv:2202.07125.

\bibitem{li2019enhancing}
S.~Li, X.~Jin, Y.~Xuan, X.~Zhou, W.~Chen, Y.-X. Wang, X.~Yan, Enhancing the
  locality and breaking the memory bottleneck of transformer on time series
  forecasting, Advances in neural information processing systems 32.

\bibitem{wu2020adversarial}
S.~Wu, X.~Xiao, Q.~Ding, P.~Zhao, Y.~Wei, J.~Huang,
  \href{https://proceedings.neurips.cc/paper/2020/file/c6b8c8d762da15fa8dbbdfb6baf9e260-Paper.pdf}{Adversarial
  sparse transformer for time series forecasting}, in: H.~Larochelle,
  M.~Ranzato, R.~Hadsell, M.~Balcan, H.~Lin (Eds.), Advances in Neural
  Information Processing Systems, Vol.~33, Curran Associates, Inc., 2020, pp.
  17105--17115.
\newline\urlprefix\url{https://proceedings.neurips.cc/paper/2020/file/c6b8c8d762da15fa8dbbdfb6baf9e260-Paper.pdf}

\bibitem{lim2021temporal}
B.~Lim, S.~{\"O}. Ar{\i}k, N.~Loeff, T.~Pfister, Temporal fusion transformers
  for interpretable multi-horizon time series forecasting, International
  Journal of Forecasting 37~(4) (2021) 1748--1764.

\bibitem{tang2021probabilistic}
B.~Tang, D.~S. Matteson, Probabilistic transformer for time series analysis,
  Advances in Neural Information Processing Systems 34 (2021) 23592--23608.

\bibitem{lin2021ssdnet}
Y.~Lin, I.~Koprinska, M.~Rana, Ssdnet: State space decomposition neural network
  for time series forecasting, in: 2021 IEEE International Conference on Data
  Mining (ICDM), IEEE, 2021, pp. 370--378.

\bibitem{zhou2021informer}
H.~Zhou, S.~Zhang, J.~Peng, S.~Zhang, J.~Li, H.~Xiong, W.~Zhang, Informer:
  Beyond efficient transformer for long sequence time-series forecasting, in:
  Proceedings of the AAAI Conference on Artificial Intelligence, Vol.~35, 2021,
  pp. 11106--11115.

\bibitem{wu2021autoformer}
H.~Wu, J.~Xu, J.~Wang, M.~Long, Autoformer: Decomposition transformers with
  auto-correlation for long-term series forecasting, Advances in Neural
  Information Processing Systems 34 (2021) 22419--22430.

\bibitem{liu2021pyraformer}
S.~Liu, H.~Yu, C.~Liao, J.~Li, W.~Lin, A.~X. Liu, S.~Dustdar, Pyraformer:
  Low-complexity pyramidal attention for long-range time series modeling and
  forecasting, in: International Conference on Learning Representations, 2021.

\bibitem{shen2022tcct}
L.~Shen, Y.~Wang,
  \href{https://www.sciencedirect.com/science/article/pii/S0925231222000571}{Tcct:
  Tightly-coupled convolutional transformer on time series forecasting},
  Neurocomputing 480 (2022) 131--145.
\newblock \href
  {http://dx.doi.org/https://doi.org/10.1016/j.neucom.2022.01.039}
  {\path{doi:https://doi.org/10.1016/j.neucom.2022.01.039}}.
\newline\urlprefix\url{https://www.sciencedirect.com/science/article/pii/S0925231222000571}

\bibitem{cirstea2022triformer}
R.-G. Cirstea, C.~Guo, B.~Yang, T.~Kieu, X.~Dong, S.~Pan,
  \href{https://doi.org/10.24963/ijcai.2022/277}{Triformer: Triangular,
  variable-specific attentions for long sequence multivariate time series
  forecasting}, in: L.~D. Raedt (Ed.), Proceedings of the Thirty-First
  International Joint Conference on Artificial Intelligence, {IJCAI-22},
  International Joint Conferences on Artificial Intelligence Organization,
  2022, pp. 1994--2001, main Track.
\newblock \href {http://dx.doi.org/10.24963/ijcai.2022/277}
  {\path{doi:10.24963/ijcai.2022/277}}.
\newline\urlprefix\url{https://doi.org/10.24963/ijcai.2022/277}

\bibitem{chen2022learning}
W.~Chen, W.~Wang, B.~Peng, Q.~Wen, T.~Zhou, L.~Sun,
  \href{https://doi.org/10.1145/3534678.3539234}{Learning to rotate: Quaternion
  transformer for complicated periodical time series forecasting}, in:
  Proceedings of the 28th ACM SIGKDD Conference on Knowledge Discovery and Data
  Mining, KDD '22, Association for Computing Machinery, New York, NY, USA,
  2022, p. 146–156.
\newblock \href {http://dx.doi.org/10.1145/3534678.3539234}
  {\path{doi:10.1145/3534678.3539234}}.
\newline\urlprefix\url{https://doi.org/10.1145/3534678.3539234}

\bibitem{drouin22tactis}
A.~Drouin, E.~Marcotte, N.~Chapados,
  \href{https://proceedings.mlr.press/v162/drouin22a.html}{{TACT}i{S}:
  Transformer-attentional copulas for time series}, in: K.~Chaudhuri,
  S.~Jegelka, L.~Song, C.~Szepesvari, G.~Niu, S.~Sabato (Eds.), Proceedings of
  the 39th International Conference on Machine Learning, Vol. 162 of
  Proceedings of Machine Learning Research, PMLR, 2022, pp. 5447--5493.
\newline\urlprefix\url{https://proceedings.mlr.press/v162/drouin22a.html}

\bibitem{zhou2022fedformer}
T.~Zhou, Z.~Ma, Q.~Wen, X.~Wang, L.~Sun, R.~Jin,
  \href{https://proceedings.mlr.press/v162/zhou22g.html}{{FED}former: Frequency
  enhanced decomposed transformer for long-term series forecasting}, in:
  K.~Chaudhuri, S.~Jegelka, L.~Song, C.~Szepesvari, G.~Niu, S.~Sabato (Eds.),
  Proceedings of the 39th International Conference on Machine Learning, Vol.
  162 of Proceedings of Machine Learning Research, PMLR, 2022, pp.
  27268--27286.
\newline\urlprefix\url{https://proceedings.mlr.press/v162/zhou22g.html}

\bibitem{liu2022non}
Y.~Liu, H.~Wu, J.~Wang, M.~Long, Non-stationary transformers: Rethinking the
  stationarity in time series forecasting, arXiv preprint arXiv:2205.14415.

\bibitem{gelper2016}
S.~Gelper, I.~Wilms, C.~Croux,
  \href{http://www.sciencedirect.com/science/article/pii/S0022435915000536}{Identifying
  demand effects in a large network of product categories}, Journal of
  Retailing 92~(1) (2016) 25 -- 39.
\newblock \href
  {http://dx.doi.org/https://doi.org/10.1016/j.jretai.2015.05.005}
  {\path{doi:https://doi.org/10.1016/j.jretai.2015.05.005}}.
\newline\urlprefix\url{http://www.sciencedirect.com/science/article/pii/S0022435915000536}

\bibitem{leeflang2008}
P.~S. Leeflang, J.~P. Selva], A.~V. Dijk], D.~R. Wittink,
  \href{http://www.sciencedirect.com/science/article/pii/S0167811608000347}{Decomposing
  the sales promotion bump accounting for cross-category effects},
  International Journal of Research in Marketing 25~(3) (2008) 201 -- 214.
\newblock \href
  {http://dx.doi.org/https://doi.org/10.1016/j.ijresmar.2008.03.003}
  {\path{doi:https://doi.org/10.1016/j.ijresmar.2008.03.003}}.
\newline\urlprefix\url{http://www.sciencedirect.com/science/article/pii/S0167811608000347}

\bibitem{srinivasan2005identifying}
S.~R. Srinivasan, S.~Ramakrishnan, S.~E. Grasman, Identifying the effects of
  cannibalization on the product portfolio, Marketing intelligence \& planning.

\bibitem{zeng2022Transformers}
A.~Zeng, M.~Chen, L.~Zhang, Q.~Xu, Are transformers effective for time series
  forecasting?, arXiv preprint arXiv:2205.13504.

\bibitem{MAKRIDAKIS20221325}
S.~Makridakis, E.~Spiliotis, V.~Assimakopoulos,
  \href{https://www.sciencedirect.com/science/article/pii/S0169207021001187}{The
  m5 competition: Background, organization, and implementation}, International
  Journal of Forecasting 38~(4) (2022) 1325--1336, special Issue: M5
  competition.
\newblock \href
  {http://dx.doi.org/https://doi.org/10.1016/j.ijforecast.2021.07.007}
  {\path{doi:https://doi.org/10.1016/j.ijforecast.2021.07.007}}.
\newline\urlprefix\url{https://www.sciencedirect.com/science/article/pii/S0169207021001187}

\bibitem{lecun1998gradient}
Y.~LeCun, L.~Bottou, Y.~Bengio, P.~Haffner, Gradient-based learning applied to
  document recognition, Proceedings of the IEEE 86~(11) (1998) 2278--2324.

\bibitem{johnson2015semi}
R.~Johnson, T.~Zhang, Semi-supervised convolutional neural networks for text
  categorization via region embedding, in: C.~Cortes, N.~Lawrence, D.~Lee,
  M.~Sugiyama, R.~Garnett (Eds.), Advances in Neural Information Processing
  Systems, Vol.~28, Curran Associates, Inc., 2015.

\bibitem{kalchbrenner2016neural}
N.~Kalchbrenner, L.~Espeholt, K.~Simonyan, A.~van~den Oord, A.~Graves,
  K.~Kavukcuoglu, \href{https://arxiv.org/abs/1610.10099}{Neural machine
  translation in linear time}, 2016.
\newline\urlprefix\url{https://arxiv.org/abs/1610.10099}

\bibitem{yin2017comparative}
W.~Yin, K.~Kann, M.~Yu, H.~Sch{\"u}tze, Comparative study of cnn and rnn for
  natural language processing, arXiv preprint arXiv:1702.01923.

\bibitem{oord2016wavenet}
A.~van~den Oord, S.~Dieleman, H.~Zen, K.~Simonyan, O.~Vinyals, A.~Graves,
  N.~Kalchbrenner, A.~Senior, K.~Kavukcuoglu,
  \href{https://arxiv.org/abs/1609.03499}{Wavenet: A generative model for raw
  audio}, in: Arxiv, 2016.
\newline\urlprefix\url{https://arxiv.org/abs/1609.03499}

\bibitem{borovykh2017conditional}
A.~Borovykh, S.~Bohte, C.~W. Oosterlee, Conditional time series forecasting
  with convolutional neural networks, arXiv preprint arXiv:1703.04691.

\bibitem{sen2019think}
R.~Sen, H.-F. Yu, I.~S. Dhillon, Think globally, act locally: A deep neural
  network approach to high-dimensional time series forecasting, Advances in
  neural information processing systems 32.

\bibitem{bai2018empirical}
S.~Bai, J.~Z. Kolter, V.~Koltun, An empirical evaluation of generic
  convolutional and recurrent networks for sequence modeling, arXiv preprint
  arXiv:1803.01271.

\bibitem{amazon}
Amazon,
  \href{https://ts.gluon.ai/api/gluonts/gluonts.model.transformer.html}{Gluonts
  transformer estimator}.
\newline\urlprefix\url{https://ts.gluon.ai/api/gluonts/gluonts.model.transformer.html}

\bibitem{Yuqietal-2023-PatchTST}
Y.~Nie, N.~H.~Nguyen, P.~Sinthong, J.~Kalagnanam, A time series is worth 64
  words: Long-term forecasting with transformers, in: International Conference
  on Learning Representations, 2023.

\bibitem{10.1137/1.9781611976700.56}
X.~Y. Xiaoyong~Jin, Yu-Xiang~Wang,
  \href{https://epubs.siam.org/doi/abs/10.1137/1.9781611976700.56}{Inter-series
  attention model for covid-19 forecasting} (2021) 495--503\href
  {http://arxiv.org/abs/https://epubs.siam.org/doi/pdf/10.1137/1.9781611976700.56}
  {\path{arXiv:https://epubs.siam.org/doi/pdf/10.1137/1.9781611976700.56}},
  \href {http://dx.doi.org/10.1137/1.9781611976700.56}
  {\path{doi:10.1137/1.9781611976700.56}}.
\newline\urlprefix\url{https://epubs.siam.org/doi/abs/10.1137/1.9781611976700.56}

\bibitem{DBLP:journals/corr/VaswaniSPUJGKP17}
A.~Vaswani, N.~Shazeer, N.~Parmar, J.~Uszkoreit, L.~Jones, A.~N. Gomez,
  L.~Kaiser, I.~Polosukhin, \href{http://arxiv.org/abs/1706.03762}{Attention is
  all you need}, CoRR abs/1706.03762.
\newblock \href {http://arxiv.org/abs/1706.03762} {\path{arXiv:1706.03762}}.
\newline\urlprefix\url{http://arxiv.org/abs/1706.03762}

\bibitem{gehring2017convolutional}
J.~Gehring, M.~Auli, D.~Grangier, D.~Yarats, Y.~N. Dauphin, Convolutional
  sequence to sequence learning (2017).
\newblock \href {http://arxiv.org/abs/1705.03122} {\path{arXiv:1705.03122}}.

\bibitem{zerveas2020Transformerbased}
G.~Zerveas, S.~Jayaraman, D.~Patel, A.~Bhamidipaty, C.~Eickhoff, A
  transformer-based framework for multivariate time series representation
  learning (2020).
\newblock \href {http://arxiv.org/abs/2010.02803} {\path{arXiv:2010.02803}}.

\bibitem{anonymous2022mqTransformer}
Anonymous, \href{https://openreview.net/forum?id=rxF4IN3R2ml}{{MQT}ransformer:
  Multi-horizon forecasts with context dependent and feedback-aware attention},
  in: Submitted to The Tenth International Conference on Learning
  Representations, 2022, under review.
\newline\urlprefix\url{https://openreview.net/forum?id=rxF4IN3R2ml}

\bibitem{makridakasAccuracy2022}
S.~Makridakis, E.~Spiliotis, V.~Assimakopoulos,
  \href{https://www.sciencedirect.com/science/article/pii/S0169207021001874}{M5
  accuracy competition: Results, findings, and conclusions}, International
  Journal of Forecasting 38~(4) (2022) 1346--1364, special Issue: M5
  competition.
\newblock \href
  {http://dx.doi.org/https://doi.org/10.1016/j.ijforecast.2021.11.013}
  {\path{doi:https://doi.org/10.1016/j.ijforecast.2021.11.013}}.
\newline\urlprefix\url{https://www.sciencedirect.com/science/article/pii/S0169207021001874}

\bibitem{Cerqueira_2020}
V.~Cerqueira, L.~Torgo, I.~Mozeti{\v{c}},
  \href{https://doi.org/10.1007%2Fs10994-020-05910-7}{Evaluating time series
  forecasting models: an empirical study on performance estimation methods},
  Machine Learning 109~(11) (2020) 1997--2028.
\newblock \href {http://dx.doi.org/10.1007/s10994-020-05910-7}
  {\path{doi:10.1007/s10994-020-05910-7}}.
\newline\urlprefix\url{https://doi.org/10.1007%2Fs10994-020-05910-7}

\end{thebibliography}

\end{document}